\documentclass[11pt]{article}

\usepackage[preprint]{acl}
\usepackage{times}
\usepackage{latexsym}

\usepackage[T1]{fontenc}

\usepackage[utf8]{inputenc}

\usepackage{microtype}

\usepackage{inconsolata}

\usepackage{graphicx}

\usepackage{amsmath}

\usepackage{booktabs}
\usepackage{multirow}
\usepackage{xcolor}
\usepackage{amssymb} 
\usepackage[linesnumbered,ruled,vlined]{algorithm2e}
\newcommand{\best}[1]{\textbf{#1}} 
\newcommand{\second}[1]{\underline{#1}} 
\usepackage{xcolor}
\usepackage[most]{tcolorbox}
\definecolor{bg_gray}{RGB}{248, 248, 248} 
\definecolor{frame_gray}{RGB}{180, 180, 180} 
\newtcolorbox{promptbox}[2][]{
    enhanced,
    breakable,             
    colback=bg_gray,       
    colframe=frame_gray,   
    boxrule=0.5pt,         
    arc=2mm,               
    left=3mm, right=3mm, top=3mm, bottom=3mm, 
    fontupper=\small\ttfamily, 
    title=\textbf{#2},     
    coltitle=black,        
    colbacktitle=bg_gray!85!black, 
    attach boxed title to top left={xshift=3mm, yshift*=-\tcboxedtitleheight/2},
    boxed title style={
        frame code={ \path[fill=bg_gray!85!black] (frame.south west) rectangle (frame.north east); }
    },
    #1
}
\usepackage{enumitem}
\usepackage{subcaption}

\usepackage{eso-pic}

\AddToShipoutPictureBG{%
  \AtTextLowerLeft{
    \raisebox{-1.7cm}{
      \parbox{\textwidth}{
        \textcolor{black}{\rule{\textwidth}{0.4pt}} \\ 
        \textcolor{black}{\textbf{Under Review}} 
      }%
    }%
  }%
}

\title{SARE: Sample-wise Adaptive Reasoning for \\
Training-free Fine-grained Visual Recognition}

\author{
Jingxiao Yang$^{1,*}$, 
DaLin He$^{1,*}$, 
Miao Pan$^{1}$, 
Kaixiang Yao$^{1}$,
Ge Su$^{1,\dagger}$, 
Wenqi Zhang$^{1}$, \\
\textbf{Yifeng Hu}$^{2}$, 
\textbf{Tangwei Li}$^{2}$, 
\textbf{Yuke Li}$^{2}$, 
\textbf{Xuhong Zhang}$^{1,\dagger}$ \\
$^{1}$Zhejiang University \\
$^{2}$Netease Yidun AI Lab \\
$^{*}$Equal contribution \quad
$^{\dagger}$Corresponding authors
}
\begin{document}
\maketitle
\begin{abstract}

Recent advances in Large Vision–Language Models (LVLMs) have enabled training-free Fine-Grained Visual Recognition (FGVR). However, effectively exploiting LVLMs for FGVR remains challenging due to the inherent visual ambiguity of subordinate-level categories. 
Existing methods predominantly adopt either retrieval-oriented or reasoning-oriented paradigms to tackle this challenge, but both are constrained by two fundamental limitations:
(1) They apply the same inference pipeline to all samples without accounting for uneven recognition difficulty, thereby leading to suboptimal accuracy and efficiency; 
(2) The lack of mechanisms to consolidate and reuse error-specific experience causes repeated failures on similar challenging cases.
To address these limitations, we propose \textbf{SARE}, 
a \underline{\textbf{S}}ample-wise \underline{\textbf{A}}daptive \underline{\textbf{RE}}asoning framework for training-free FGVR.
Specifically, SARE adopts a cascaded design that combines fast candidate retrieval with fine-grained reasoning, invoking the latter only when necessary. In the reasoning process, SARE incorporates a self-reflective experience mechanism that leverages past failures to provide transferable discriminative guidance during inference, without any parameter updates.
Extensive experiments across 14 datasets substantiate that SARE achieves state-of-the-art performance while substantially reducing computational overhead. 

\end{abstract}

\section{Introduction}
\label{Introduction}

\begin{figure}[t]
    \centering
    \begin{subfigure}[b]{1.0\linewidth}
        \centering
        \includegraphics[width=\linewidth]{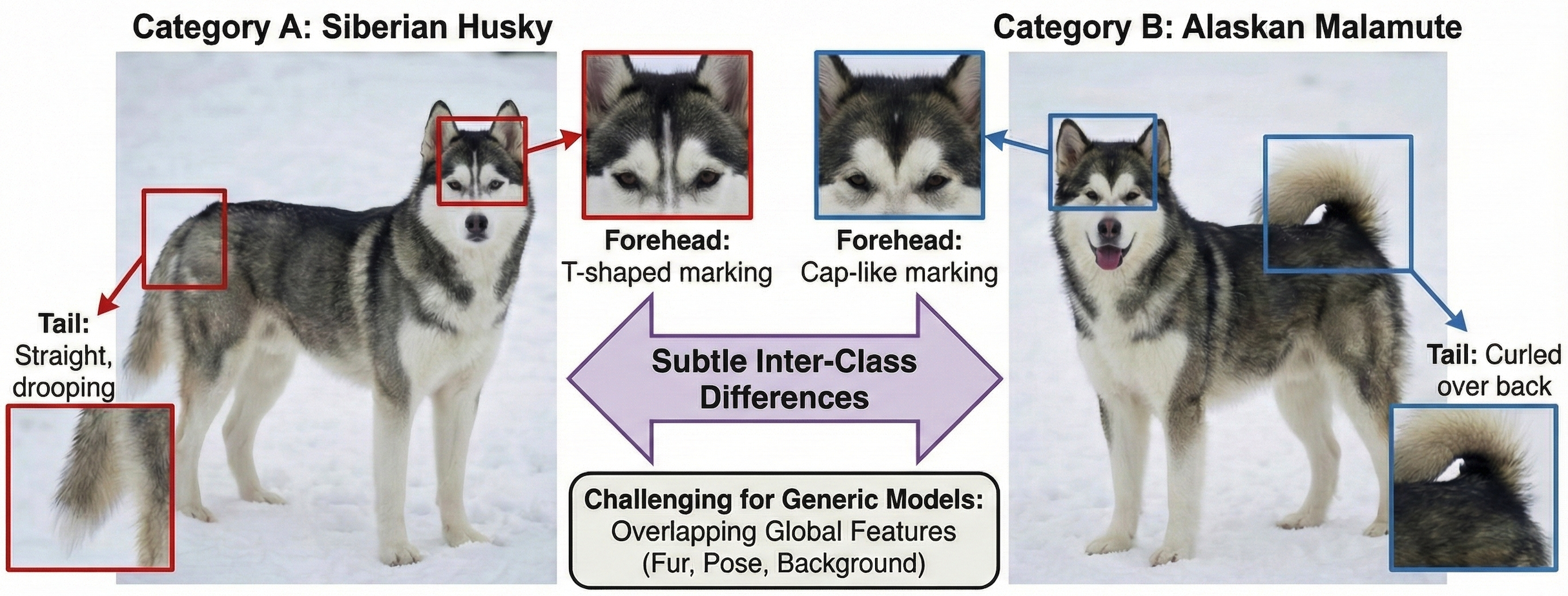}
        \caption{The challenge of FGVR.}
        \label{fig:example_a}
    \end{subfigure}
    
    \begin{subfigure}[b]{1.0\linewidth}
        \centering
        \includegraphics[width=\linewidth]{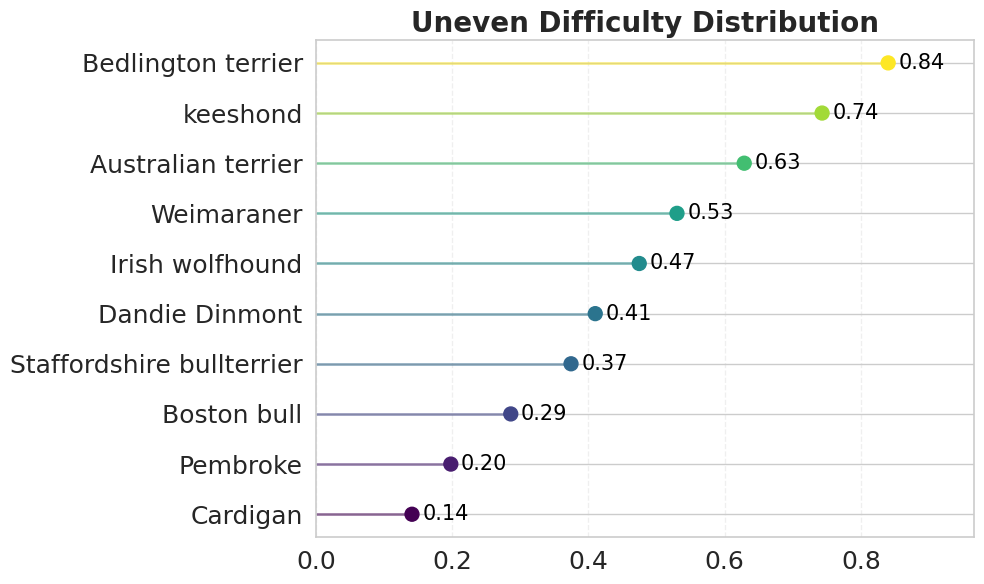}
        \caption{CLIP normalized scores corresponds to a 70\% Top-1 Acc.}
        \label{fig:example_b}
    \end{subfigure}
    
    \caption{(a) Fine-Grained Visual Recognition challenge: sub-categories (e.g., Siberian Husky vs. Alaskan Malamute) exhibit subtle visual differences, requiring attention to localized discriminative features such as forehead patterns and tail shape. (b) For a fixed overall retrieval accuracy (70\%), different sub-categories require substantially different confidence thresholds, indicating that identical confidence scores imply varying reliability across categories.}
    \label{fig:motivation_combined}
\end{figure}

Recent advances in Large Vision-Language Models (LVLMs) have led to remarkable progress across a wide range of visual recognition tasks~\cite{qwen,qwen3,InternVL3.5,deepseek,gemini2.5,openai2023gpt4}. 
However, despite their strong general capabilities, LVLMs remain weak on Fine-Grained Visual Recognition (FGVR)~\cite{performence1,analysis1,peng2026survey,FineDefics}.
FGVR aims to distinguish visually similar sub-categories, such as different bird species or dog breeds~\cite{intro1,Kim_2024}. Unlike general visual recognition where coarse visual features often suffice, FGVR presents a unique challenge: decisive and discriminative cues are subtle, highly localized, and easily confounded~\cite{intro1,peng2026survey}. As shown in Figure~\ref{fig:example_a}, two dog breeds may only differ in forehead pattern and tail shape. This challenge is further exacerbated by large intra-class variations caused by changes in pose, background, etc.

In light of these challenges, a variety of methods have been proposed to improve model's performance in FGVR. These approaches can be broadly categorized into two paradigms, according to how they employ LVLMs: i) \textbf{Retrieval-oriented methods} prompt the LVLM to generate fine-grained textual descriptions and perform multimodal matching. These methods are efficient and scalable, but rely primarily on global representations and struggle to capture localized discriminative cues required to distinguish visually similar sub-categories~\cite{is-CLIP,FG-CLIP}. ii) \textbf{Reasoning-oriented methods} instead formulate recognition as a multi-choice visual question-answering (VQA)~\cite{vqa} problem by presenting candidate labels as context. While capable of deeper analysis, as the candidate space grows, the model's attention becomes increasingly diffused, leading to unstable reasoning behaviors such as hallucination and performance degradation~\cite{Cascade,whyvlm,wu2025languageoverrulesrevealingtext,FineDefics}.

In spite of different strategies, both paradigms share two fundamental limitations. First, they apply the same inference pipeline to all samples regardless of their difficulty. As shown in Figure~\ref{fig:example_b}, even with a 70\% top-1 retrieval accuracy, the confidence scores vary significantly across categories, indicating that sample difficulty is highly uneven. Applying uniform processing wastes computation on easy cases while providing insufficient analysis for hard ones. Moreover, for simple samples, excessive reasoning may introduce overthinking and hallucination~\cite{hallucination,COUNTS}, actually degrading performance. Second, existing methods are designed to be stateless: unlike human experts who learn from past mistakes to refine their judgment criteria, these methods cannot accumulate experience across instances, leading to repeated mistakes in similar scenarios.


To address the aforementioned limitations, we draw inspiration from how humans learn and handle complex visual tasks. On the one hand, Humans rely on a dual-system mechanism~\cite{surveyreasoning} to cope with varying recognition difficulty: System 1 performs rapid, intuitive judgments based on salient cues and is effective for clear, easily distinguishable cases, whereas System 2 engages in deliberate, nuanced reasoning to resolve ambiguous or highly similar examples. On the other hand, humans continuously distill experience from errors, forming transferable decision criteria that prevent repeated mistakes in similar situations.

Building on these insights, we propose \textbf{SARE}, 
a \underline{\textbf{S}}ample-wise \underline{\textbf{A}}daptive \underline{\textbf{RE}}asoning framework designed for training-free FGVR.
First, we conceptualize the image-text retrieval and matching process as \textbf{System 1}:
it operates rapidly and intuitively, leveraging global visual-semantic alignment to handle straightforward samples efficiently. 
Subsequently, we formulate the LVLM-based nuanced reasoning process as \textbf{System 2}: it engages in deliberate, step-by-step analysis to resolve samples where subtle inter-class differences demand focused attention on discriminative details.

The synergy between the dual-system is crucial. System 1 solely cannot handle hard samples with subtle visual differences, while System 2 applied uniformly to all samples not only incurs prohibitive computational costs but also introduces overthinking on easy cases, potentially causing hallucination and degraded accuracy. To this end, SARE introduces a statistics-based trigger that dynamically determines whether System 2 reasoning is required for each sample. This trigger accounts for three factors: the model's confidence on the current sample, the historical difficulty of its category, and the ambiguity among candidate options, thereby routing each sample to the appropriate processing depth.

To support both efficient retrieval and robust reasoning, we construct three lightweight knowledge bases from a small set with labeled category names: 
(1) A multimodal prototype library containing textual and visual prototypes for fast retrieval; (2) A statistical retrieval library that records category-level retrieval statistics, enabling category-aware and robust trigger decisions; (3) An experience library derived from model self-reflection to guide reasoning in challenging scenarios. 
When System 2 is activated, it receives the Top-$K_c$ candidates from System 1 along with guidance from the experience library, enabling the LVLM to focus on truly discriminative fine-grained cues and behave more like a domain expert.
In a nutshell, main contributions are summarized as:

\begin{itemize}
    \item 
    We propose SARE, a sample-wise adaptive reasoning framework for training-free FGVR. SARE departs from uniform inference by dynamically adapting its inference strategy to sample difficulty, improving efficiency and accuracy  without any parameter updates.

    \item SARE effectively synergizes fast retrieval (System 1) with nuanced reasoning (System 2) via a dynamic trigger mechanism. Furthermore, it incorporates a self-reflective experience module that learns from past errors to distill reusable discriminative guidance, enabling the system to continuously improve through accumulated experience.

    \item Extensive experiments across 14 datasets demonstrate the superiority of SARE.Notably, it outperforms leading training-free methods by over 8\% and training-based baselines by 1.64\%. Meanwhile, SARE also exhibits superior robustness on out-of-distribution tasks and reduces computational overhead.
\end{itemize}

\section{Methodology}
\label{methodology}

\begin{figure*}[ht]
    \centering
    \includegraphics[width=\textwidth]{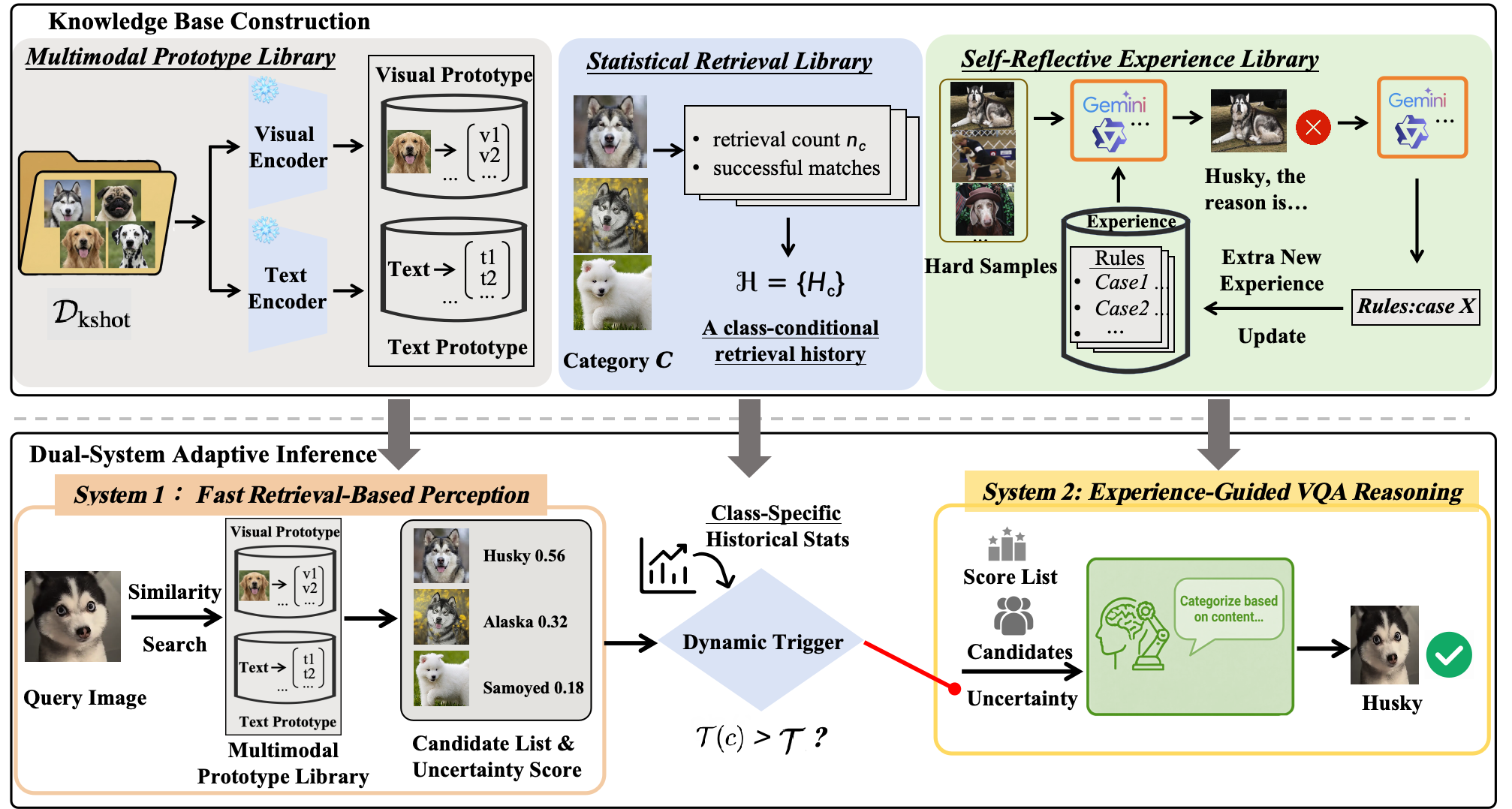} 
    \caption{\textbf{Overview of SARE}. The framework performs fast prototype-based retrieval to generate candidate categories for a query image, followed by a class-conditional trigger that adaptively invokes fine-grained reasoning when retrieval confidence is insufficient. Experience distilled from past errors is injected as contextual guidance, enabling accurate and efficient training-free FGVR.}
    \label{fig:overview}
\end{figure*}


We present SARE, a sample-wise adaptive reasoning framework tailored to training-free FGVR.
SARE mainly addresses two fundamental issues identified in prior work: the highly uneven difficulty of fine-grained instances and the stateless inference paradigm that fails to leverage past errors. As illustrated in Figure~\ref{fig:overview}, SARE explicitly decouples fast perception from nuanced reasoning. Each sample is first processed by a lightweight retrieval module(System 1), and a statistics-driven trigger adaptively determines whether additional reasoning is required. For ambiguous cases, SARE invokes experience-guided reasoning(System 2), where reusable discriminative experience distilled from historical failures is injected to support reliable fine-grained decisions.

\subsection{Knowledge Base Construction}
\label{sec:3.1}
To support sample-wise adaptive inference without parameter updates, SARE constructs lightweight and reusable knowledge offline from a $k_{shot}$ sub-category support set $\mathcal{D}_{\text{kshot}}$. This knowledge provides how perception, uncertainty estimation, and reasoning are performed at inference time. Specifically, we build three complementary knowledge bases with non-overlapping roles.

\noindent \textbf{Multimodal Prototype Library.}
\label{sec:Prototype}
Fast perception requires a compact and stable representation that supports reliable early decisions. Purely visual similarity is often unstable under large intra-class variations, while textual semantics provide complementary global structure. We therefore construct multimodal prototypes to support efficient retrieval.

Specifically, for category $c \in D_{kshot}$, we compute: (i) a visual prototype by averaging CLIP image embeddings; (ii) a textual prototype encoded from an LVLM-generated fine-grained description of the category.
These prototypes form the visual prototype set $P_v$ and textual prototype set $P_t$, respectively. They are then used exclusively by the fast retrieval module (System 1) without any additional model overhead.

\noindent \textbf{Statistical Retrieval Library.}
\label{sec:Statistical}
While System 1 produces similarity-based confidence scores, their reliability varies significantly across sub-categories~\cite{misupervisedlearning}, rendering static thresholds unsuitable (Figure~\ref{fig:example_b}). To calibrate these variations, we construct a statistical retrieval library that captures class-conditional retrieval behavior. We perform retrieval on ${D}_{\text{kshot}}$ using the multimodal prototypes $P_v$ and $P_t$. For category $c$, we record its retrieval count $n_c$ and the number of successful matches $N$, forming a class-conditional retrieval history. These statistics serve as class-conditional priors for uncertainty estimation and are used by the dynamic trigger in Formula~\ref{eq:threshold}.

\noindent \textbf{Self-Reflective Experience Library.}
To mitigate repeated errors on challenging cases, we introduce a self-reflective experience library that consolidates past errors into reusable discriminative guidance.
We formalize each inference on the ${D_{kshot}}$ as a trajectory:
\begin{equation}
    \mathcal{T} = (I, \mathcal{C}, \tau, \hat{y}, y),
    \label{eq:trajectory} 
\end{equation}
where $I$ denotes the query image, $\mathcal{C}$ is the retrieved candidate set, $\tau$ is the reasoning path, and $\hat{y}$ and $y$ are the predicted and ground-truth labels, respectively.
When an error occurs (i.e., $\hat{y} \neq y$), a retrospective analysis is triggered to identify overlooked discriminative cues from $\tau$. These cues are abstracted into compact, structured decision rules and stored as experience entries $\mathcal{E}=\{e_i\}$ .
To keep the experience library compact and informative, we apply a lightweight maintenance policy that merges semantically complementary rules, filters redundant ones, and replaces low-value entries with more informative experience. During inference, relevant experience entries are retrieved as contextual guidance for fine-grained reasoning, enabling the model to avoid repeating similar mistakes without any parameter updates. For details, please refer to the Appendix~\ref{sec:appendix_self_ref}.

\subsection{Sample-wise Adaptive Inference}
\label{sec:3.2}

FGVR datasets exhibit highly uneven sample difficulty: many instances can be resolved using coarse visual cues, while others require careful inspection of subtle, localized attributes. Applying a uniform pipeline is therefore structurally inefficient and risks either wasting computation on easy samples or under-analyzing hard ones.
SARE adopts a sample-wise adaptive inference mechanism, inspired by the observation that LVLMs often achieve low Top-$1$ but high Top-$K_c$ accuracy on fine-grained tasks~\cite{Scan}. This suggests that retrieval can reliably narrow down candidates, while reasoning should be reserved for genuinely ambiguous cases. This observation motivates a clear separation between fast perceptual filtering and deliberate fine-grained reasoning.

\noindent \textbf{System 1: Fast Retrieval-based Perception.}
System 1 is a lightweight perception module that performs nearest-neighbor matching between the query image and the multimodal prototypes $P_v$ and $P_t$. It involves only similarity computation and incurs minimal computational overhead.

System 1 serves two purposes: (1) it enables early exit for samples with clear decision boundaries; and (2) it produces a compact Top-$K_c$ candidate set $\mathcal{C} = \{c_1, \ldots, c_K\}$ along with similarity-based confidence scores. These outputs provide the necessary signals for the subsequent trigger to assess reliability and decide whether further reasoning is required.

\noindent \textbf{System 2: Experience-guided Nuanced Reasoning.}
\label{sec:system2}
System 2 is activated only for ambiguous cases with low retrieval confidence and is formulated as a VQA inference.
Specifically, System 2 takes the input image $I$, the Top-$K_c$ candidate set $\mathcal{C}$, and a small set of retrieved experience entries $\mathcal{E}$ as contextual input to the LVLM. The prediction is formalized as:
\begin{equation}
    \hat{y} = f(I, \mathcal{C}, \mathcal{E}).
\end{equation}
By constraining the candidate space and injecting structured experience, System 2 is guided to attend to truly discriminative fine-grained cues, enabling precise reasoning in similar ambiguous scenarios without resorting to repeated model invocations or costly multi-step prompting.
We provide the complete prompt templates and implementation details in Appendix~\ref{sec:appendix_prompts} to ensure reproducibility.

\subsection{Statistics-based Dynamic Trigger}

The dynamic trigger serves as the decision-making core of SARE, managing the transition from System 1 to System 2. Its primary objective is to selectively allocate expensive computational resources to genuinely ambiguous cases while preserving the efficiency of rapid perception for simple samples.

Given the Top-$K_c$ retrieved category candidates from System 1, our goal is to assess whether the top-$1$ prediction is sufficiently reliable. In practice, System 1 may fail under three common conditions: 
(1) calibration mismatch between visual and textual similarities; 
(2) category-dependent confidence bias from VLM embeddings (Figure~\ref{fig:example_b});
(3) high ambiguity among candidates.
To address cross-modal calibration mismatch, we first normalize visual and textual similarities using temperature-scaled Softmax to obtain distributions $P_{img}$ and $P_{text}$, which are linearly fused as:
\begin{equation}
P_{\mathrm{fuse}} = \lambda P_{img} + (1-\lambda) P_{text},
\end{equation}
where $\lambda$ is set to 0.3. 
Since probability values can still be poorly calibrated across modalities, we further incorporate Reciprocal Rank Fusion (RRF)~\cite{RRF} to exploit robust rank-level agreement. 
For each category $c \in C$, we compute an RRF score based on its rank $r_m(c)$:
\begin{equation}
    \mathcal{R}(c) = \sum_{m \in \{v, t\}} \frac{1}{\kappa + r_m(c)},
\end{equation}
where $r_m(c)$ denotes the rank of category $c$ within modality $m$, $\kappa=60$ is a smoothing constant. This metric is robust to score outliers as it relies solely on rank positions.
The final retrieval confidence for category $c$ is defined as
\begin{equation}
\hat{p}_c = P_{\mathrm{fuse}}(c) + \beta \cdot \mathcal{R}(c),
\end{equation}
where $\beta=0.1$. Based on the fused confidence, we define a trigger score for the Top-$1$ candidate as
\begin{equation}
\mathcal{G}(c)=\hat{p}_c-\eta \sqrt{\frac{\log N}{2 n_c}}-\alpha H(\mathbf{p}_c),
\label{eq:threshold}
\end{equation}
where each term captures a distinct source of uncertainty in System 1.
Specifically, $\hat{p}_c$ reflects the fused retrieval confidence. The second term, derived from Hoeffding's inequality (Sec.~\ref{sec:Statistical}), penalizes categories with limited historical retrieval evidence ($n_c$), preventing overconfident decisions on statistically unreliable candidates. The third term $H(\mathbf{p}_c)$ measures ambiguity among the Top-$K_c$ candidates: high entropy indicates that System 1 fails to clearly separate competing hypotheses, even when absolute confidence is high.

If $\mathcal{G}(c)$ exceeds a predefined threshold $\theta$, the prediction from System 1 is accepted; otherwise, System 2 is activated for fine-grained reasoning. 
This statistics-driven trigger allows SARE to selectively allocate computation to ambiguous cases, effectively decoupling inference cost from recognition difficulty.

\section{Experiments}
\label{experiments}


\begin{table*}[t]
\centering
\resizebox{\textwidth}{!}{%
\begin{tabular}{l|ccccccc|ccccc|c|c}
\toprule
\multirow{2}{*}{\textbf{Method}} & \multicolumn{7}{c|}{\textbf{Fine-Grained Visual Recognition (FGVC)}} & \multicolumn{5}{c|}{\textbf{General Recognition}} & \multirow{2}{*}{\textbf{Avg.}} & \multirow{2}{*}{\textbf{$\Delta$}} \\
\cmidrule(lr){2-8} \cmidrule(lr){9-13}
 & Dogs & Flowers & CUB & Cars & Pets & Air. & Bird. & Food & DTD & IN-1k & SUN & UCF & & \\
\midrule
\multicolumn{15}{c}{\textit{\textbf{Backbone}}} \\
\midrule
CLIP-B/32 & 52.09 & 63.44 & 53.26 & 62.38 & 81.66 & 23.67 & 42.37 & 80.32 & 44.26 & 63.67 & 62.32 & 64.13 & 57.80 & -2.80 \\
Qwen2.5-VL-7B & 65.90 & 68.50 & 43.70 & 75.31 & 85.12 & 54.01 & 54.48 & 69.90 & 68.83 & 68.40 & 58.11 & 73.30 & 65.46 & +4.86 \\
\midrule
\multicolumn{15}{c}{\textit{\textbf{Training-free LVLM Methods}}} \\
\midrule
\multicolumn{15}{l}{\textit{(Retrieval-based Methods)}} \\
SuS-X-LC & 62.35 & 72.17 & 55.02 & 66.32 & 86.89 & 27.96 & 46.83 & 84.32 & 52.34 & 66.87 & 68.02 & 66.22 & 62.94 & +2.34 \\
FineR & 52.88 & 67.67 & 56.75 & 65.36 & 83.24 & 24.96 & 43.69 & 81.31 & 49.76 & 68.65 & 65.72 & 67.26 & 60.60 & +0.00 \\
E-FineR & 52.72 & 68.96 & 57.89 & 65.89 & 82.96 & 25.66 & 44.32 & 81.56 & 50.27 & 69.02 & 66.96 & 67.74 & 61.16 & +0.56 \\
RAR & 53.29 & 67.99 & 57.01 & 66.01 & 84.37 & 24.39 & 43.87 & 81.65 & 49.57 & 70.26 & 67.09 & 67.01 & 61.04 & +0.44 \\
AWT & 68.35 & 76.13 & 59.54 & 70.03 & 90.76 & 30.93 & 48.63 & 85.58 & 55.20 & 71.32 & 70.33 & 69.12 & 66.33 & +5.73 \\
ProtoMM & 68.63 & 77.40 & 60.04 & 69.92 & \second{91.29} & 31.02 & 48.79 & 85.89 & 56.38 & 72.76 & 70.67 & 70.62 & 66.95 & +6.35 \\
SCAN & 69.98 & 77.86 & 64.83 & 77.96 & 89.78 & 39.68 & 53.57 & 86.37 & 62.65 & 72.14 & 72.83 & 73.71 & 70.11 & +9.51 \\
\midrule
\multicolumn{15}{l}{\textit{(Reasoning-based Methods)}} \\
AutoSEP & 67.82 & 70.63 & 63.19 & 68.12 & 86.49 & 58.76 & 60.34 & 84.75 & 61.47 & 71.45 & 71.92 & 67.53 & 69.37 & +8.77 \\
CascadeVLM & 64.54 & 74.23 & 60.26 & 79.26 & 86.17 & 29.18 & 56.27 & 84.51 & 53.24 & 73.98 & 69.96 & 69.05 & 66.72 & +6.12 \\
UniFGVC & 65.23 & \best{95.84} & 78.80 & \second{94.60} & 90.92 & 61.10 & 66.25 & 82.30 & 73.90 & 81.10 & 76.60 & 80.90 & 78.96 & +18.36 \\
\midrule
\multicolumn{15}{c}{\textit{\textbf{Training-based LVLM Methods}}} \\
\midrule
FineDefics & 73.36 & 87.81 & 55.46 & 85.01 & 84.73 & 64.08 & 45.72 & 84.26 & 63.53 & 71.41 & 68.90 & 75.97 & 71.69 & +11.09 \\
VT-FSL & \best{84.63} & \second{93.64} & \best{90.98} & 92.75 & 88.92 & \second{80.31} & \second{76.57} & \best{91.32} & \second{76.31} & \best{89.75} & \best{81.98} & \second{85.37} & \second{86.04} & \second{+25.44} \\
\midrule
\textbf{Ours (SARE)} & \second{84.29} & 88.31 & \second{90.76} & \best{99.34} & \best{95.38} & \best{83.02} & \best{87.30} & \second{88.02} & \best{83.10} & \second{85.06} & \second{80.05} & \best{87.58} & \best{87.68} & \best{+27.08} \\
\bottomrule
\end{tabular}%
}
\caption{Comparison with state-of-the-art methods on 7 fine-grained and 5 general datasets. We highlight the \best{best} results in bold and the \second{second-best} results with underlining. ``Avg.'' denotes the average accuracy across all 12 datasets. $\Delta$ indicates the performance gain compared to FineR.}
\label{tab:main_results}
\end{table*}

\subsection{Setup}

\noindent \textbf{Datasets.} 
We carefully select 14 datasets to systematically assess its fine-grained discrimination ability, general recognition performance, and robustness under distribution shifts:
(1) Fine-grained recognition (7 datasets): We select widely-used benchmarks that cover diverse domains and subtle category distinctions, including CUB-200-2011~\cite{wah2011caltech}, Stanford Dogs~\cite{KhoslaYaoJayadevaprakashFeiFei_FGVC2011}, Stanford Cars~\cite{krause20133d}, FGVC-Aircraft~\cite{maji2013fine}, Oxford-IIIT Pet~\cite{parkhi2012cats}, Oxford 102 Flowers~\cite{nilsback2008automated} and Birdsnap~\cite{berg2014birdsnap}.
(2) General recognition (5 datasets) Beyond fine-grained tasks, we evaluate general visual recognition on  Food-101~\cite{bossard2014food}, ImageNet-1K~\cite{deng2009imagenet}, Describable Textures Dataset (DTD)~\cite{cimpoi2014describing}, SUN397~\cite{Xiao2014SUNDE}, and UCF101~\cite{soomro2012ucf101}.
(3) Robustness evaluation (2 datasets): To probe performance under distribution shifts, we incorporate challenging ImageNet variants: ImageNet-V2~\cite{recht2019imagenet}, and ImageNet-Sketch~\cite{wang2019learning}.
Detailed dataset descriptions are provided in Appendix~\ref{sec:appendix_dataset}.

\noindent \textbf{Baseline.}
We compare SARE with several representative SoTA FGVR methods, selected to ensure fair and comprehensive coverage of both training-free and training-based paradigms.
(i) Training-free methods contain retrieval-based baselines, such as Sus-X-LC~\cite{sus}, FineR~\cite{FineR}, E-FineR~\cite{E-FineR}, AWT~\cite{AWT}, ProtoMM~\cite{ProtoMM}, CascadeVLM~\cite{Cascade} and SCAN~\cite{Scan}, as well as reasoning-oriented frameworks including UniFGVR~\cite{UniFGVR}, AutoSEP~\cite{AutoSEP}, and MCQA~\cite{AttentionIntervention}.
(ii) Training-based methods are also selected as reference baselines, including two of the most representative cross-modal alignment methods, i.e., ternary contrastive learning FineDefics~\cite{FineDefics} and volume alignment VT-FSL~\cite{VT-FSL}. 

\noindent \textbf{Implementation details.}
We use Qwen2.5-VL-7B~\cite{qwen} as the LVLM backbone for textual prototype generation, experience construction, and System 2 reasoning. For System 1, CLIP-B/32~\cite{CLIP} encodes visual inputs and textual prototypes for multimodal prototype construction and candidate retrieval.
We set the number of candidate categories $K_c=10$, experience $\mathcal{E}=8$ and $k_{shot}=3$. We emphasize that all methods’ settings are strictly consistent.
All experiments were conducted on a single NVIDIA RTX A800 GPU with fixed random seeds.

\subsection{Main Results}
 \textbf{Best overall performance.} As shown in Table~\ref{tab:main_results},  SARE reaches an average accuracy of 87.68\%, substantially outperforming all training-free baselines and surpassing strong training-based methods such as FineDefics by 15.88\% and VT-FSL by 1.64\%. Unlike prior approaches that exhibit pronounced performance fluctuations across domains (e.g., UniFGVC), SARE consistently exhibits high performance on both fine-grained and general datasets, indicating robust cross-domain generalization.
The improvements are particularly notable on highly ambiguous fine-grained datasets (e.g., Aircraft, Birdsnap and Dogs), where recognition relies on subtle inter-class differences. These gains underscore the advantage of experience-guided reasoning, as the distilled experience help the model focus on truly decisive attributes.
To further analyze performance under varying difficulty levels, we selecte two datasets based on their intrinsic difficulty (Table~\ref{tab:trigger_rates}): {Stanford Cars} (relatively easy) and {Stanford Dogs} (challenging).
As illustrated in Figure~\ref{fig:efficiency_car} and Figure~\ref{fig:efficiency_dog} (In Appendix), SARE not only improves accuracy but also achieves a superior accuracy–efficiency trade-off, dynamically allocating more computation to challenging samples while reducing effort for easier ones.

\section{Analysis}
\label{sec:analysis}

\subsection{Effectiveness of Dynamic Trigger}
\begin{figure}[t]
    \centering
    \includegraphics[width=1.0\linewidth]{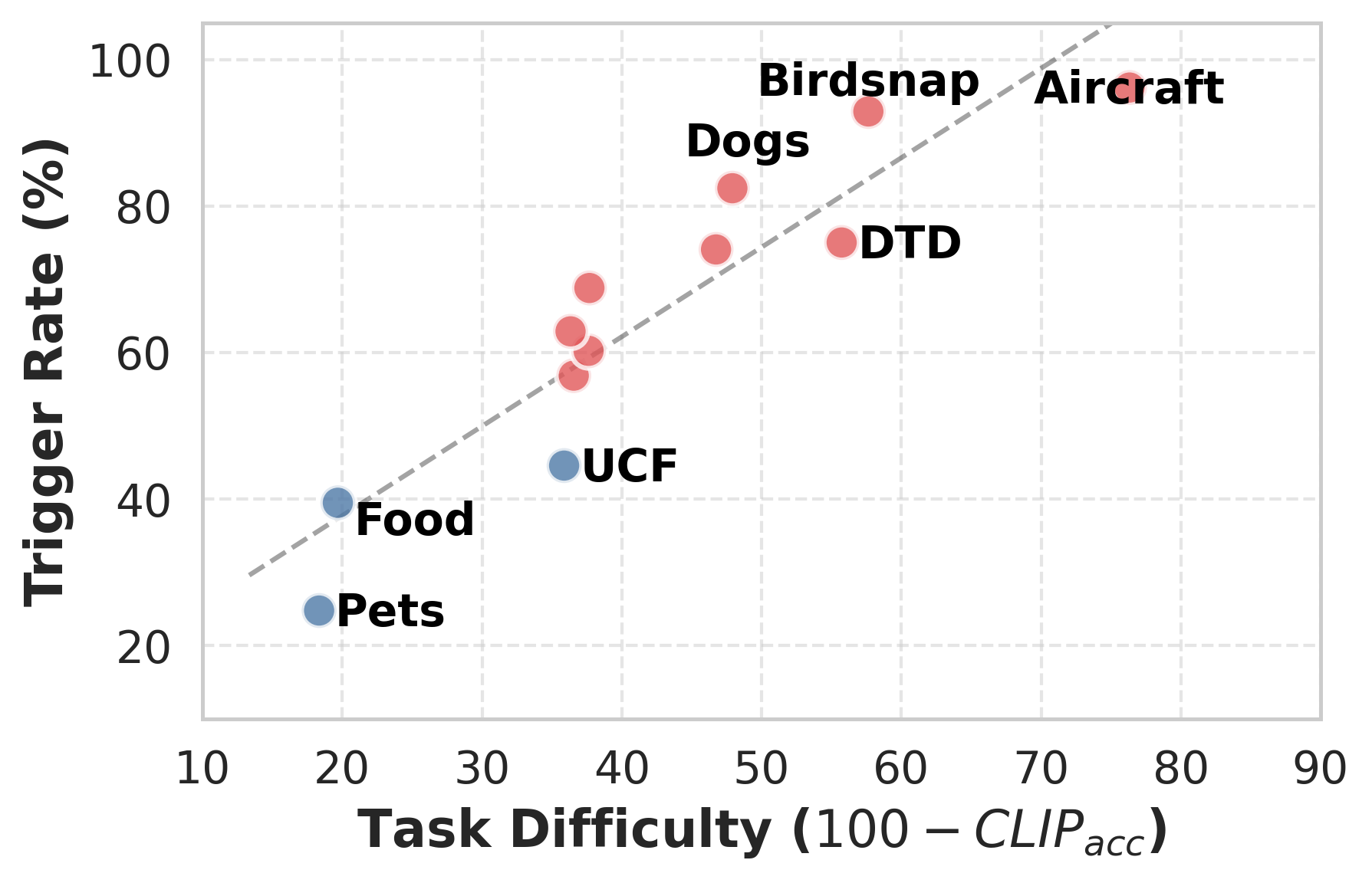} 
    \caption{The proportion of samples triggering System 2 across datasets with varying recognition difficulty. The x-axis measures dataset-level difficulty using $100\%-CLIP_{Top-1}$ accuracy.}
    \label{fig:trigger_analysis}
\end{figure}

\textbf{Efficient difficulty adaptation.} As shown in Figure~\ref{fig:trigger_analysis}, we analyze the performance of the dynamic trigger. SARE dynamically adjusts the activation rate of System 2 according to sample difficulty. When recognition is easy, the trigger rate remains low. As difficulty increases, the trigger rate rises accordingly, allocating more samples to fine-grained reasoning. This strong correlation indicates that the dynamic trigger effectively learns \emph{when} additional reasoning is necessary, rather than blindly increasing computation. Table~\ref{tab:system1_reliability} shows that samples identified as requiring only System 1 achieve very high accuracy, demonstrating that the trigger maintains both efficiency and reliable recognition.

\begin{figure}[t]
    \centering
    \includegraphics[width=1.0\linewidth]{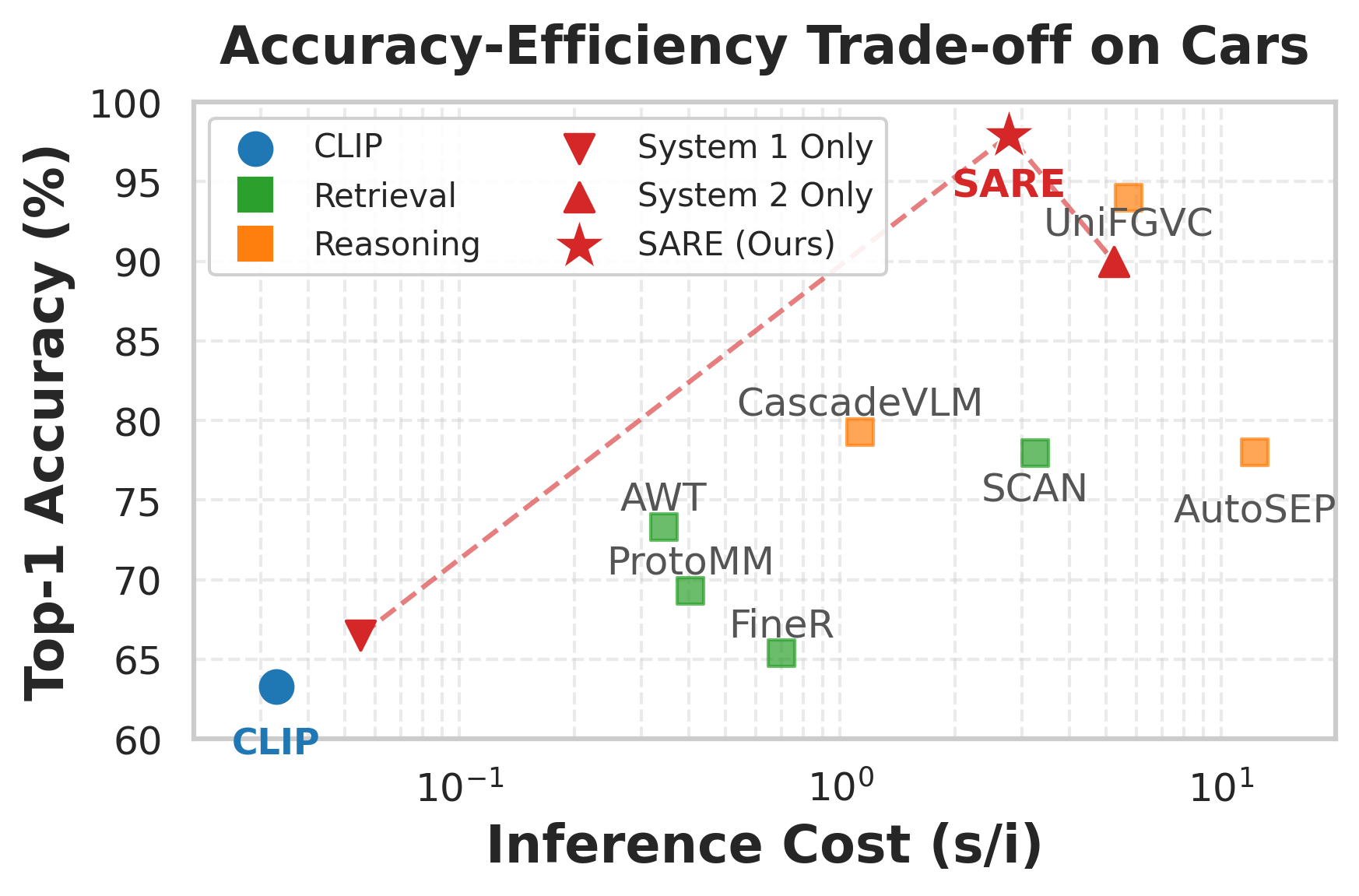} 
    \caption{Comparison of SARE against baselines on Stanford Cars dataset. SARE achieves the optimal balance, significantly outperforming baselines in accuracy with lower inference overhead.} 
    \label{fig:efficiency_car}
\end{figure}

\begin{table}[h]
\centering
\small
\begin{tabular}{ccc|cc}
\toprule
\textbf{System 1} & \textbf{System 2} & \textbf{Experience} & \textbf{Dogs} & \textbf{CUB} \\ 
\midrule
\checkmark &  & & 58.24 & 73.02 \\
& \checkmark & & 78.61 & 78.43 \\
\checkmark & \checkmark & & 76.01 & 81.36\\
& \checkmark & \checkmark & 81.34 & 88.21 \\
\checkmark & \checkmark & \checkmark & \textbf{84.29} & \textbf{90.76} \\
\bottomrule
\end{tabular}
\caption{Ablation study of key components in SARE. 
}
\label{tab:ablation_component}
\end{table}


\begin{table}[h]
\centering
\resizebox{\linewidth}{!}{
\begin{tabular}{l|cc|cc}
\toprule
\multirow{2}{*}{\textbf{Method}} & \multicolumn{2}{c|}{\textbf{ImageNet-V2}} & \multicolumn{2}{c}{\textbf{ImageNet-Sketch}} \\
& Acc. (\%) & $\Delta$ & Acc. (\%) & $\Delta$ \\
\midrule
\textit{$SARE_{tgt}$} & {84.43} & {-} & {72.49} & {-} \\
\textbf{$SARE_{src}$} & {83.72} & \textbf{-0.69} & {71.96} & \textbf{-0.53} \\
\bottomrule
\end{tabular}
}
\caption{Comparison of SARE with experience constructed on ImageNet-1K ($SARE_{src}$) versus experience constructed on target-domain experience ($SARE_{tgt}$).}
\label{tab:generalization}
\end{table}


\subsection{Ablation Studies}
\noindent \textbf{Component Effectiveness.}
As shown in Table~\ref{tab:ablation_component}, relying solely on retrieval-based System 1 yields an accuracy of 58.24\%, indicating that global semantic alignment struggles with subtle inter-class differences. Incorporating reasoning-based System 2 boosts performance to 76.65\%. This confirms that VQA-based reasoning captures fine-grained details via attribute localization. Finally, the experience library brings a further gain, reaching the best accuracy of 84.29\%. This demonstrates that experience act as effective {in-context demonstrations}, guiding the LVLM to discriminative regions and enhanced fine-grained recognition capability.

\noindent \textbf{Hyperparameter Analysis.}
In Appendix~\ref{sec:appendix_hyperparameter}, we analyze SARE’s sensitivity to three key hyperparameters: the candidate set size $K_c$, the number of retrieved experience entries $\mathcal{E}$, and the $k$-shot size of $D_{kshot}$. As shown in Figures~\ref{fig:hyperparameter} and Figure~\ref{fig:k_shot_sensitivity}, SARE remains stable across a wide range of values, achieving the best performance at $K_c=10$ and $\mathcal{E}=8$, and exhibiting strong robustness to variations in $k_{shot}$.
In Appendix~\ref{sec:appendix_sensitivity}, we evaluate the impact of different sampling strategies for $k_{shot}$ in $D_{kshot}$, The results show that SARE maintains stable performance across six distinct strategies.


\begin{figure}[h]
    \centering
    \begin{subfigure}[b]{0.48\textwidth}
        \includegraphics[width=\textwidth]{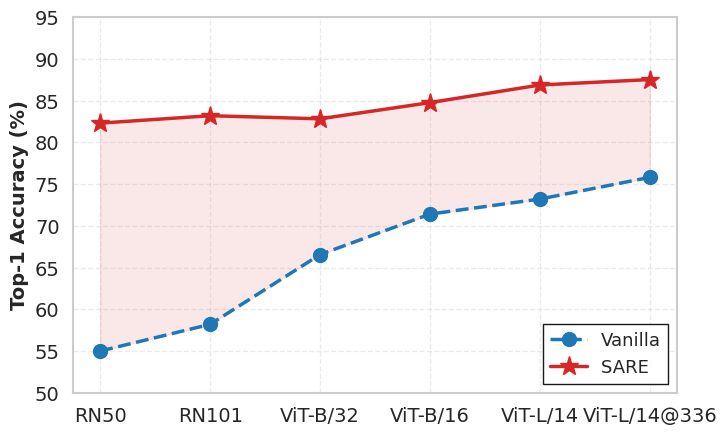} 
        \caption{VLM Backbone Performance}
        \label{fig:clip_backbone}
    \end{subfigure}
    \hfill
    \begin{subfigure}[b]{0.48\textwidth}
        \includegraphics[width=\textwidth]{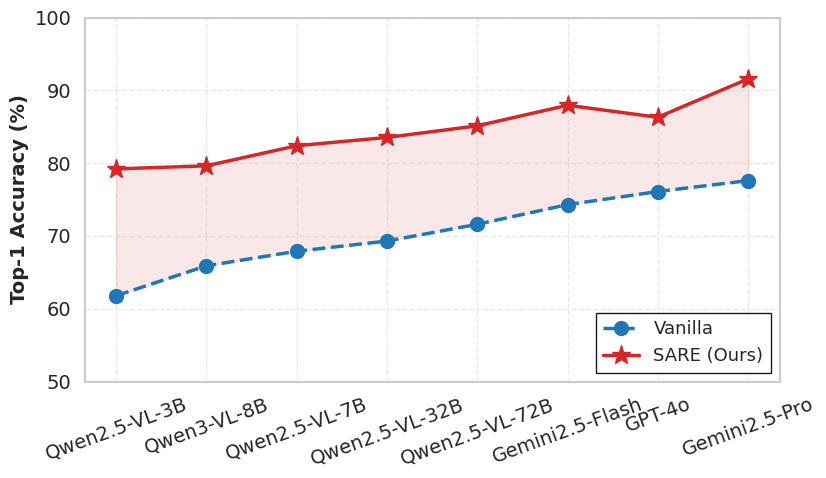} 
        \caption{LVLM Backbone Performance}
        \label{fig:lvlm_backbone}
    \end{subfigure}
    \caption{Performance of SARE across different backbone architectures.
SARE consistently enhances performance across all backbones, with larger relative gains on lower-performing models.}
    \label{fig:backbone_comparison}
\end{figure}

\subsection{Performance on different Backbones}
\label{sec:backone}
\textbf{Strong model-agnostic robustness.}
To assess the model-agnostic property of SARE, we conduct controlled experiments by separately varying the visual backbone (VLM) and the reasoning backbone (LVLM). Specifically, when evaluating different VLMs, we fix the LVLM as Qwen2.5-VL-7B; when evaluating different LVLMs, we fix the visual encoder as CLIP ViT-B/32.
As shown in Figure~\ref{fig:clip_backbone} and Figure~\ref{fig:lvlm_backbone}, SARE consistently delivers strong performance across all evaluated backbones, demonstrating its robustness and model-agnostic applicability. Across different VLMs, SARE yields stable performance improvements regardless of backbone capacity. Notably, the relative gains are more pronounced for weaker visual backbones (e.g., RN50 and RN101) and gradually diminish as the base representations become stronger. This trend indicates that SARE effectively compensates for limited perceptual capacity rather than relying on architecture-specific characteristics. A similar pattern is observed for LVLMs. Although larger models achieve higher absolute accuracy, SARE consistently enhances performance across different model scales, with more substantial gains on smaller LVLMs.
Overall, these results indicate that SARE provides complementary reasoning benefits rather than relying on backbone-specific characteristics, demonstrating strong robustness and model-agnostic applicability.

\begin{figure}[t]
    \centering
    \includegraphics[width=1.0\linewidth]{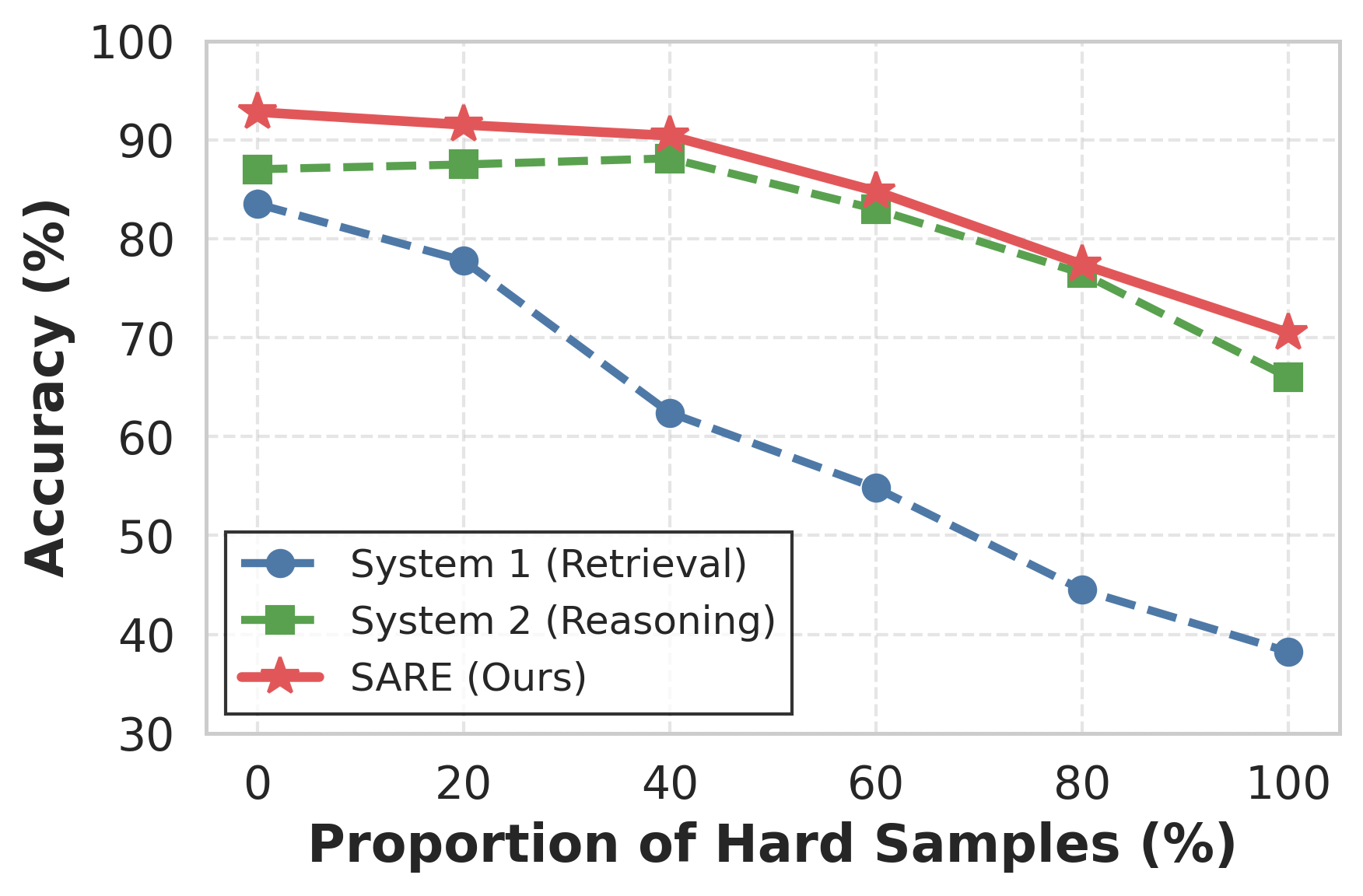} 
    \caption{Behavior of different inference strategies under increasing sample difficulty on Stanford Dogs.}
    \label{fig:behavior}
\end{figure}

\subsection{Transferability Analysis}
\textbf{Experience is transferable.} To examine this, we perform cross-domain evaluations on ImageNet-V2 and ImageNet-Sketch. We build the experience library on ImageNet-1K and directly apply it to the target domains, denoted as $SARE_{src}$, and compare it with a target-domain upper bound where the experience library is constructed directly from the target dataset, denoted as $SARE_{tgt}$.
As shown in Table~\ref{tab:generalization}, $SARE_{src}$ achieves performance comparable to $SARE_{tgt}$, with only marginal accuracy degradation, even under significant distribution and style shifts. This indicates that the learned experience captures transferable discriminative cues, enabling the model to resolve a family of similar challenging cases rather than overfitting to domain-specific visual patterns.

\subsection{Behavioral Analysis}
\textbf{Mitigating overthinking \& underthinking.} To analyze how different inference strategies behave under varying sample difficulty, we perform a controlled study on the Stanford Dogs dataset by splitting it into \emph{easy} and \emph{hard} subsets based on retrieval correctness, and then creating five test sets with gradually increasing proportions of hard samples, from 0\% to 100\%.  
As shown in Figure~\ref{fig:behavior}, System 1’s overall performance steadily declines with higher difficulty, reflecting its limited ability to handle highly ambiguous instances. System 2, by contrast, shows a counterintuitive accuracy increase when hard samples constitute from 0 to 40\% of the test set, which we attribute to overreasoning on simpler examples that can sometimes degrade performance. SARE, however, consistently achieves the highest accuracy across all difficulty levels, demonstrating its effectiveness in avoiding overthinking on easy samples while effectively addressing the challenging ones.


\subsection{Qualitative Analysis and Case Studies}
A qualitative case study illustrating how experience is constructed and used to guide model reasoning is presented in Appendix~\ref{sec:appendix_case_study}. Figure~\ref{fig:heatmap} provides a heatmap highlighting how SARE influences and guides the model’s decisions.

\section{Conclusion}
\label{sec:conclusion}
We present SARE, a sample-wise adaptive reasoning framework for training-free FGVR.
It adaptively integrates fast retrieval (System 1) with fine-grained VQA-style reasoning (System 2) via a sample-wise trigger. The trigger invokes reasoning only when retrieval is unreliable. A self-reflective experience mechanism abstracts past errors into transferable discriminative rules, guiding reasoning toward key discriminative cues without any parameter updates. 
SARE consistently achieves SoTA results across 14 datasets while reducing time costs, and the rationality of SARE's design has been verified through extensive experiments.

\clearpage
\section*{Limitations}
Despite demonstrating state-of-the-art performance and efficiency in training-free FGVR, we emphasize that SARE has some limitations.
First, although deriving transferable guidance from past errors is a promising design, further improving the quality of this experience would require additional self-reflection, inevitably introducing extra computational overhead.
Second, while the trigger mechanism is effective, SARE adopts a conservative triggering strategy for very challenging datasets (e.g., Stanford Dogs, FGVC-Aircraft), resulting in a higher-than-expected activation rate. We plan to explore more accurate and efficient dynamic triggers in future work.
Finally, SARE primarily enhances LVLMs from a textual context perspective, future work could explore complementary strategies that leverage visual features to further improve fine-grained recognition.

\bibliography{acl_latex}

\clearpage
\appendix
\section*{Appendix}

The appendix is organized as follows:

\begin{itemize}
    \item {Section~\ref{sec:appendix_datasets}} summarizes statistics and descriptions of the 14 datasets used in our experiments.
    
    \item {Section~\ref{sec:appendix_prompts}} details the prompt templates for Knowledge Base construction, System 2 reasoning, and the Self-Reflective mechanism.
    
    \item {Section~\ref{sec:appendix_hyperparameter}} presents analysis of key hyperparameters, including candidate count ($K_c$), experience capacity ($\mathcal{E}$), and $k_{shot}$ size.

    \item {Section~\ref{sec:appendix_sensitivity}} evaluates the effect of different $k$-shot support set sampling strategies on performance.
    
    \item {Section~\ref{sec:appendix_case_study}} provides qualitative case studies illustrating how the Experience Library guides model reasoning.
    \item {Section~\ref{sec:appendix_related_work}} provides related work for this work.
\end{itemize}

\section{Experimental Setup}
\label{sec:appendix_datasets}

\begin{table*}[t]
\centering
\resizebox{\linewidth}{!}{
\begin{tabular}{lcccccccccccc}
\toprule
\textbf{Dataset} & \textbf{Aircraft} & \textbf{Birdsnap} & \textbf{Dogs} & \textbf{DTD} & \textbf{CUB} & \textbf{SUN} & \textbf{IN-1K} & \textbf{Cars} & \textbf{Flowers} & \textbf{UCF} & \textbf{Food} & \textbf{Pets} \\
\midrule
\textbf{Trigger Rate (\%)} & 96.19 & 92.92 & 82.44 & 75.03 & 74.08 & 68.82 & 62.90 & 60.24 & 56.85 & 44.57 & 39.49 & 24.78 \\
\bottomrule
\end{tabular}
}
\caption{\textbf{Analysis of Trigger Rates.} The percentage of samples activating the System 2 varies significantly based on dataset difficulty.}
\label{tab:trigger_rates}
\end{table*}


\begin{table*}[t]
\centering

\resizebox{\linewidth}{!}{
\begin{tabular}{lcccccccccccc}
\toprule
\textbf{Dataset} & \textbf{Aircraft} & \textbf{Birdsnap} & \textbf{Dogs} & \textbf{DTD} & \textbf{CUB} & \textbf{SUN} & \textbf{IN-1K} & \textbf{Cars} & \textbf{Flowers} & \textbf{UCF} & \textbf{Food} & \textbf{Pets} \\
\midrule
\textbf{Sys-1 Acc. (\%)} & 88.93 & 96.96 & 91.52 & 100 & 95.78 & 91.98 & 99.97 & 99.37 & 97.94 & 97.74 & 92.81 & 95.43 \\
\bottomrule
\end{tabular}
}
\caption{Reliability of System 1 Decisions. We report the recognition accuracy on the subset of samples that the dynamic trigger determined \emph{did not} require System 2 reasoning. The consistently high accuracy across datasets demonstrates that our trigger effectively identifies easy samples where retrieval-based perception is sufficient.}
\label{tab:system1_reliability}
\end{table*}

\subsection{Dataset Details}
\label{sec:appendix_dataset}
We follow the few-shot experimental protocol established in FineR. We utilize the official test split of each dataset for performance evaluation to report the results. Instead of using the full training set, we construct the few-shot support set (used for Knowledge Base construction in SARE) by randomly sampling $K$ images per category from the official training split. In our main experiments, we set the shot number $K=3$ and ensure strict data separation where $\mathcal{D}_{\text{support}} \cap \mathcal{D}_{\text{test}} = \emptyset$.
Table \ref{tab:appendix_datasets} summarizes the detailed statistics of the 14 datasets used in our experiments, including the number of categories and the size of the official test sets used for evaluation. For FGVC benchmarks (e.g., Stanford Dogs), we follow the standard splits adopted in prior work, and results on UCF101 are reported on Split-1.

\begin{table}[h]
\centering
\renewcommand{\arraystretch}{1.15} 
\setlength{\tabcolsep}{2.5pt} 

\resizebox{\linewidth}{!}{
    \begin{tabular}{l|ccc}
    \toprule
    \textbf{Dataset} & \textbf{\# Cls.} & \textbf{\# Test Size} & \textbf{Domain} \\
    \midrule
    \multicolumn{4}{l}{\textit{\textbf{Fine-Grained Visual Recognition}}} \\
    \midrule
    CUB-200-2011 & 200 & 5,794 & Birds \\
    Stanford Dogs & 120 & 8,580 & Dogs \\
    Stanford Cars & 196 & 8,041 & Cars \\
    FGVC-Aircraft & 100 & 3,333 & Aircraft \\
    Oxford-IIIT Pet & 37 & 3,669 & Animals \\
    Oxford 102 Flowers & 102 & 6,149 & Flowers \\
    Birdsnap & 500 & 2,443 & Birds \\
    \midrule
    \multicolumn{4}{l}{\textit{\textbf{General Visual Recognition}}} \\
    \midrule
    Food-101 & 101 & 25,250 & Food \\
    ImageNet-1K & 1k & 50,000 & Objects \\
    DTD (Textures) & 47 & 1,880 & Textures \\
    SUN397 & 397 & 19,850 & Scenes \\
    UCF101 & 101 & 3,783 & Actions \\
    \midrule
    \multicolumn{4}{l}{\textit{\textbf{Out-of-Domain Datasets}}} \\
    \midrule
    ImageNet-V2 & 1k & 10,000 & Objects \\
    ImageNet-Sketch & 1k & 50,889 & Sketches \\
    \bottomrule
    \end{tabular}
}
\caption{Detailed statistics of the 14 datasets used in our experiments. To save space, \# Cls. denotes the number of categories.}
\label{tab:appendix_datasets}
\end{table}

\section{Prompt Templates}
\label{sec:appendix_prompts}

\subsection{Knowledge Base and Reasoning Prompt}
This section details the prompt templates used for (1) generating the initial multimodal knowledge base  and (2) performing the final System 2 adaptive reasoning. Note that the prompts for the \textit{Self-Reflective Experience Construction} are detailed separately in the subsequent Appendix~\ref{sec:appendix_self_ref} due to their iterative nature.

\begin{promptbox}[attach boxed title to top center={yshift*=-\tcboxedtitleheight/2}]{Textual Prototype}
You are an expert taxonomist specializing in fine-grained visual recognition.

\textbf{Input:} \\
$\bullet$ Category: \texttt{\{category\_name\}} \\
$\bullet$ Reference: [Set of Support Images]

\textbf{Task:} Generate a comprehensive and discriminative description that captures the key visual characteristics that distinguish this category from other similar categories.

\textbf{Focus on:}
\begin{enumerate}[leftmargin=1.5em, nosep]
    \item Distinctive physical features
    \item Color patterns and markings
    \item Size and proportions
    \item Behavioral characteristics (if applicable)
    \item Unique identifying traits
\end{enumerate}

\textbf{Constraint:} The description should be concise but informative, suitable for fine-grained visual recognition task.
\end{promptbox}

\begin{promptbox}[attach boxed title to top center={yshift*=-\tcboxedtitleheight/2}]{System 2 Inference}
You are a fine-grained recognition expert. Your task is to identify the specific sub-category of the provided image.

\textbf{Context Provided:} \\
\textbf{1.Candidate Classes} (highly likely to contain the correct option): \\
\texttt{\{candidate\_text\}}

\textbf{2. Expert Guidance} (Retrieved Experience): \\
\texttt{\{experience\_context\}}

\textbf{Task:} Please analyze the image step by step and provide:
\begin{enumerate}[leftmargin=1.5em, nosep]
    \item Your reasoning chain (Chain-of-Thought) based on the visual evidence and expert guidance.
    \item Your final prediction (only the category name).
\end{enumerate}

\textbf{Output Format:} \\
Reasoning: [your step-by-step reasoning] \\
Prediction: [category name]
\end{promptbox}

\subsection{The implementation of Self-Reflection}
\label{sec:appendix_self_ref}
The construction of the Experience Library is not a single-step generation but a closed-loop process involving reasoning, analysis, and strategy update. We detail the specific prompts for each phase below.

\begin{promptbox}[attach boxed title to top center={yshift*=-\tcboxedtitleheight/2}]{Step 1: Initial Self-Belief Reasoning}
You are an expert in fine-grained visual recognition. Please follow these steps to identify the object:

\begin{enumerate}[leftmargin=1.5em, nosep]
    \item \textbf{Observe:} Look at the overall object and identify its coarse category.
    \item \textbf{Localize:} Identify the most discriminative local parts.
    \item \textbf{Compare:} Recall visual characteristics of candidate subcategories.
    \item \textbf{Decide:} Choose the most likely class based on the details.
\end{enumerate}

\textbf{Constraint:} Answer ONLY with the final class name.
\end{promptbox}

If the prediction is incorrect ($\hat{y} \neq y$), the specific failure diagnosis is triggered:

\begin{promptbox}[attach boxed title to top center={yshift*=-\tcboxedtitleheight/2}]{Step 2: Specific Failure Diagnosis}
You are an expert in fine-grained visual recognition. Analyze this specific failure case where the model incorrectly predicted `\texttt{\{predicted\_category\}}' but the correct answer is `\texttt{\{true\_category\}}'.

\textbf{Context:} \\
$\bullet$ Model's Reasoning: \texttt{\{model\_reasoning\}} \\
$\bullet$ Top Candidates: \texttt{\{candidates\_info\}} \\
$\bullet$ Definition (\texttt{\{true\_category\}}): \texttt{\{correct\_category\_desc\}} \\
$\bullet$ Definition (\texttt{\{predicted\_category\}}): \texttt{\{predicted\_category\_desc\}}

\textbf{Task:} Focus ONLY on the visual evidence in this image.
\begin{enumerate}[leftmargin=1.5em, nosep]
    \item Locate the specific region where the visual feature contradicts the prediction.
    \item Compare this feature against the category definitions provided.
    \item Identify the exact visual attribute (e.g., tail shape, beak color) that caused confusion.
\end{enumerate}

\textbf{Constraint:} Do not generalize yet. Provide a detailed diagnosis of this specific image instance. Output format: \textbf{Visual Evidence} and \textbf{Direct Cause}.
\end{promptbox}

Next, the system distills this specific diagnosis into a generalized rule:

\begin{promptbox}{Step 3: Abstraction \& Generalization}
You are a knowledge engineer. Your task is to distill a specific failure diagnosis into a universal, abstract rule to guide future predictions.

\textbf{Input Data:} \\
$\bullet$ Conflict: \texttt{\{true\_category\}} vs. \texttt{\{predicted\_category\}} \\
$\bullet$ Diagnosis: \texttt{\{step2\_diagnosis\_output\}}

\textbf{Task:} Formulate a concise, high-level verification rule.
\begin{enumerate}[leftmargin=1.5em, nosep]
    \item \textbf{Abstract:} Remove references to "this image". Focus on the concept.
    \item \textbf{Actionable:} The rule should be a direct instruction for what to check.
    \item \textbf{Discriminative:} Clearly distinguish the two categories.
\end{enumerate}

\textbf{Constraint:} Return \textbf{ONLY} the rule text (under 30 words).

\textbf{Example Output:} ``To distinguish Husky from Malamute, check the tail curvature: Husky tails are straight, while Malamute tails curl over the back.''
\end{promptbox}

Finally, the model integrates the new rule to update its internal strategy:

\begin{promptbox}[attach boxed title to top center={yshift*=-\tcboxedtitleheight/2}]{Step 4: Self-Belief Strategy Update}
Based on the failure analysis and new insights, update the Self-Belief strategy.

\textbf{Input:} \\
$\bullet$ Current Strategy: \texttt{\{current\_self\_belief\}} \\
$\bullet$ New Rule/Insight: \texttt{\{failure\_analysis\}}

\textbf{Task:} Update the strategy to:
\begin{enumerate}[leftmargin=1.5em, nosep]
    \item Maintain the core recognition framework.
    \item Add specific guidance for handling similar difficult cases.
    \item Emphasize discriminative features that were previously overlooked.
\end{enumerate}

\textbf{Constraint:} Provide \textbf{only} the updated Self-Belief strategy without additional explanation.
\end{promptbox}

\begin{figure}[t]
    \centering
    \includegraphics[width=1.0\linewidth]{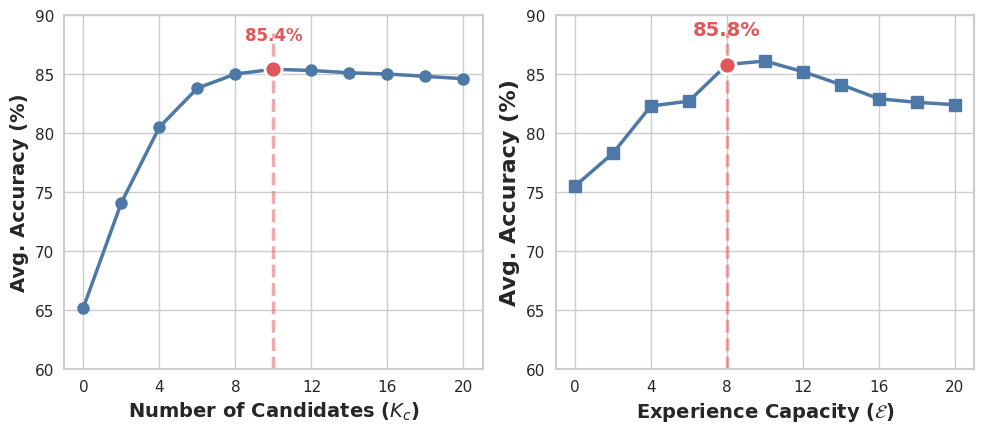}
    \caption{\textbf{Hyperparameter Sensitivity Analysis.} 
    We analyze the impact of candidate count ($K_c$) and experience capacity ($\mathcal{E}$) on average accuracy across seven fine-grained benchmarks. 
    The \textbf{line plots} reveal a clear inverted-U trend for both parameters: performance improves as context increases, peaks at the optimal settings ($K_c=10, \mathcal{E}=8$), and slightly declines beyond this point due to diminishing marginal utility and the introduction of semantic noise.}
    \label{fig:hyperparameter}
\end{figure}

\begin{figure}[t]
    \centering
    \includegraphics[width=1.0\linewidth]{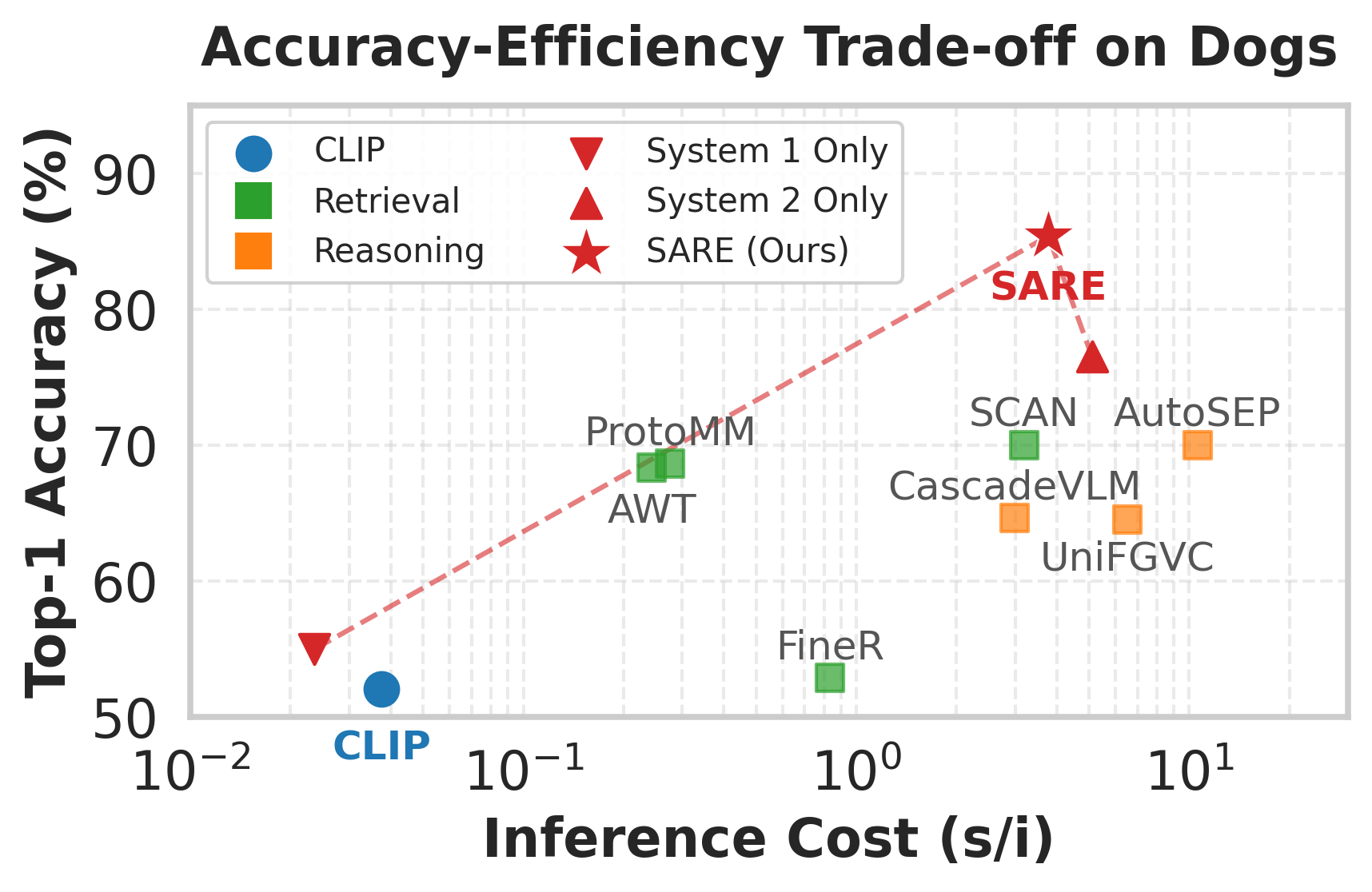} 
    \caption{Comparison of SARE against baselines on Stanford Dogs dataset.} 
    \label{fig:efficiency_dog}
\end{figure}

\section{Hyperparameter Analysis}
\label{sec:appendix_hyperparameter}
\begin{figure}[t]
    \centering

    \includegraphics[width=\linewidth]{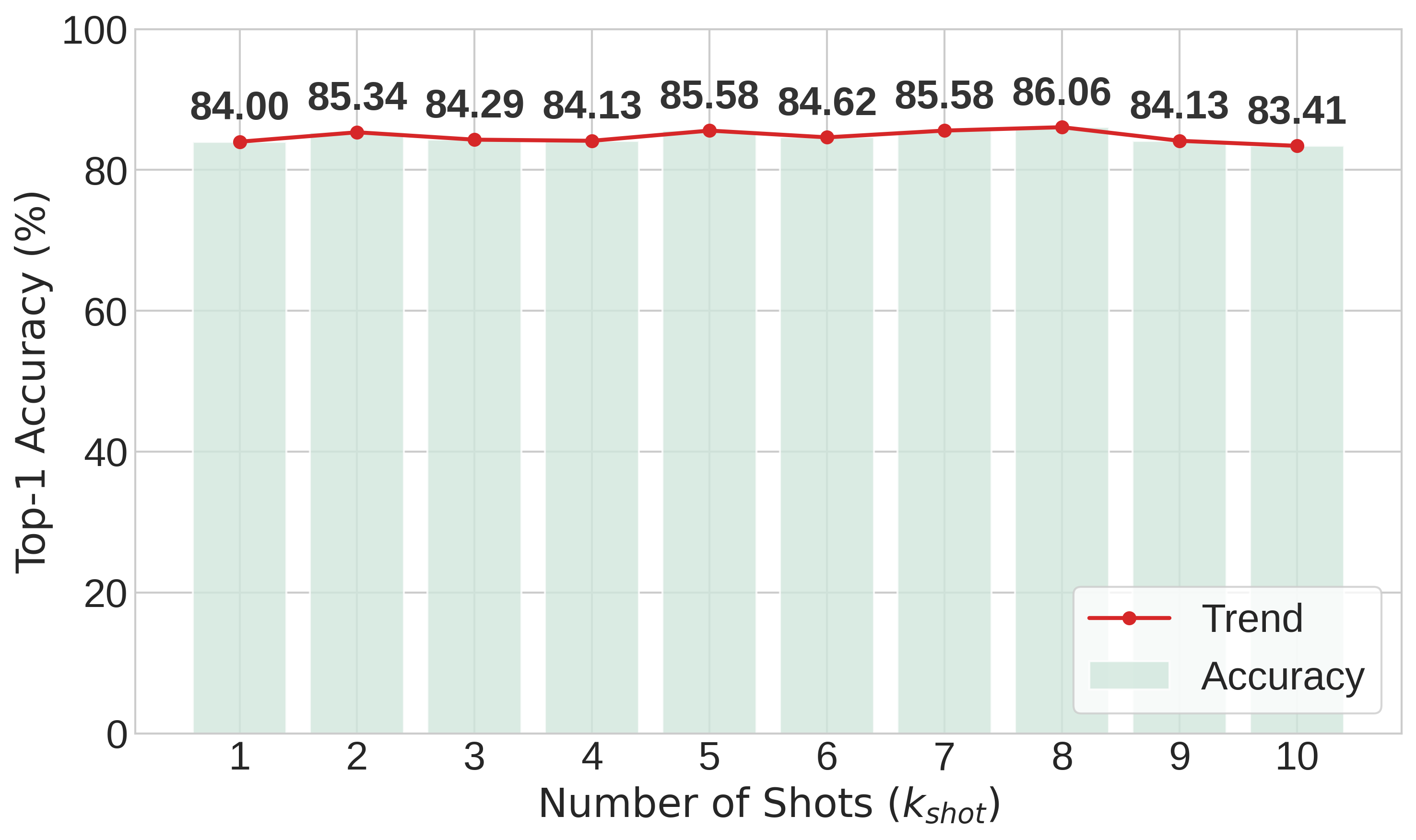}
    \caption{Sensitivity Analysis of Shot Number $k_{shot}$. The results demonstrate that SARE maintains robust performance across different few-shot settings. As shown in the figure, the accuracy fluctuation is minimal as $k$ varies from 1 to 10, indicating that our method is not overly sensitive to the exact size of the support set.}
    \label{fig:k_shot_sensitivity}
\end{figure}

In this section, we investigate the sensitivity of SARE to three critical hyperparameters: the number of retrieved candidates ($K_c$), the capacity of the Experience Library ($\mathcal{E}$) and the number of small labeled set ($k_{shot}$). We evaluate the accuracy across Stanford Dogs~\cite{KhoslaYaoJayadevaprakashFeiFei_FGVC2011}.
As illustrated in Figure~\ref{fig:hyperparameter}, when $K_c$ is small, the candidate set is unlikely to contain the correct label, which directly limits the upper bound of recognition accuracy. As $K_c$ increases, performance improves and reaches its peak around $K_c=10$, where the candidate set already includes the correct category with high probability. Further increasing $K_c$ does not yield additional gains, as the inclusion of more candidates no longer improves coverage but instead introduces redundant alternatives. 
Similarly, when the experience capacity $\mathcal{E}$ is too small, System~2 lacks sufficient contextual and historical guidance to effectively correct fine-grained reasoning errors. In contrast, excessively large values of $\mathcal{E}$ introduce irrelevant semantic noise, which disperses the LVLM’s attention and leads to a degradation in performance.
Figure~\ref{fig:k_shot_sensitivity} demonstrate that SARE exhibits remarkable robustness to the size of the support set. Specifically, the accuracy fluctuates within a narrow margin (83\%--86\%) across the range of $k_{shot}=1$ to $k_{shot}=10$. This indicates that our framework can effectively distill discriminative knowledge even from minimal examples, without being overly sensitive to the exact number of reference images. This verified its practicability and stability in scenarios with few samples

\begin{table}[h]
\centering
\small

\resizebox{0.8\linewidth}{!}{
\begin{tabular}{l|c}
\toprule
\textbf{Sampling Strategy} & \textbf{Top-1 Acc (\%)} \\
\midrule
Random     & 84.29 \\
Diversity  & 84.59 \\
Confusion  & 83.56 \\
Centroid   & 85.15 \\
Entropy    & 85.44 \\
Boundary   & 85.68 \\
\bottomrule
\end{tabular}
}
\caption{\textbf{Impact of Sampling Strategies.} Performance of SARE on Stanford Dogs under different $k$-shot sampling strategies.}
\label{tab:sampling_sensitivity}
\end{table}

\section{Sensitivity Analysis}
\label{sec:appendix_sensitivity}
To evaluate the sensitivity of SARE to the selection of the $k_{shot}$ of $D_{kshot}$, we conduct experiments on Stanford Dogs using six sampling strategies: Random, Centroid, Boundary, Entropy, Diversity, and Confusion.

As shown in Table~\ref{tab:sampling_sensitivity}, SARE exhibits stable performance across all sampling strategies, with accuracy varying within a narrow range (approximately 2\%), indicating low sensitivity to the specific composition of the support set. Notably, strategies that prioritize informative or difficult samples (Boundary and Entropy) achieve slightly higher accuracy than representative or random sampling.
This suggests that while SARE is robust to different sampling choices, constructing the Experience Library from more challenging samples can provide more effective supervisory signals for fine-grained reasoning.

\begin{figure*}[h]
    \centering
    \includegraphics[width=\linewidth]{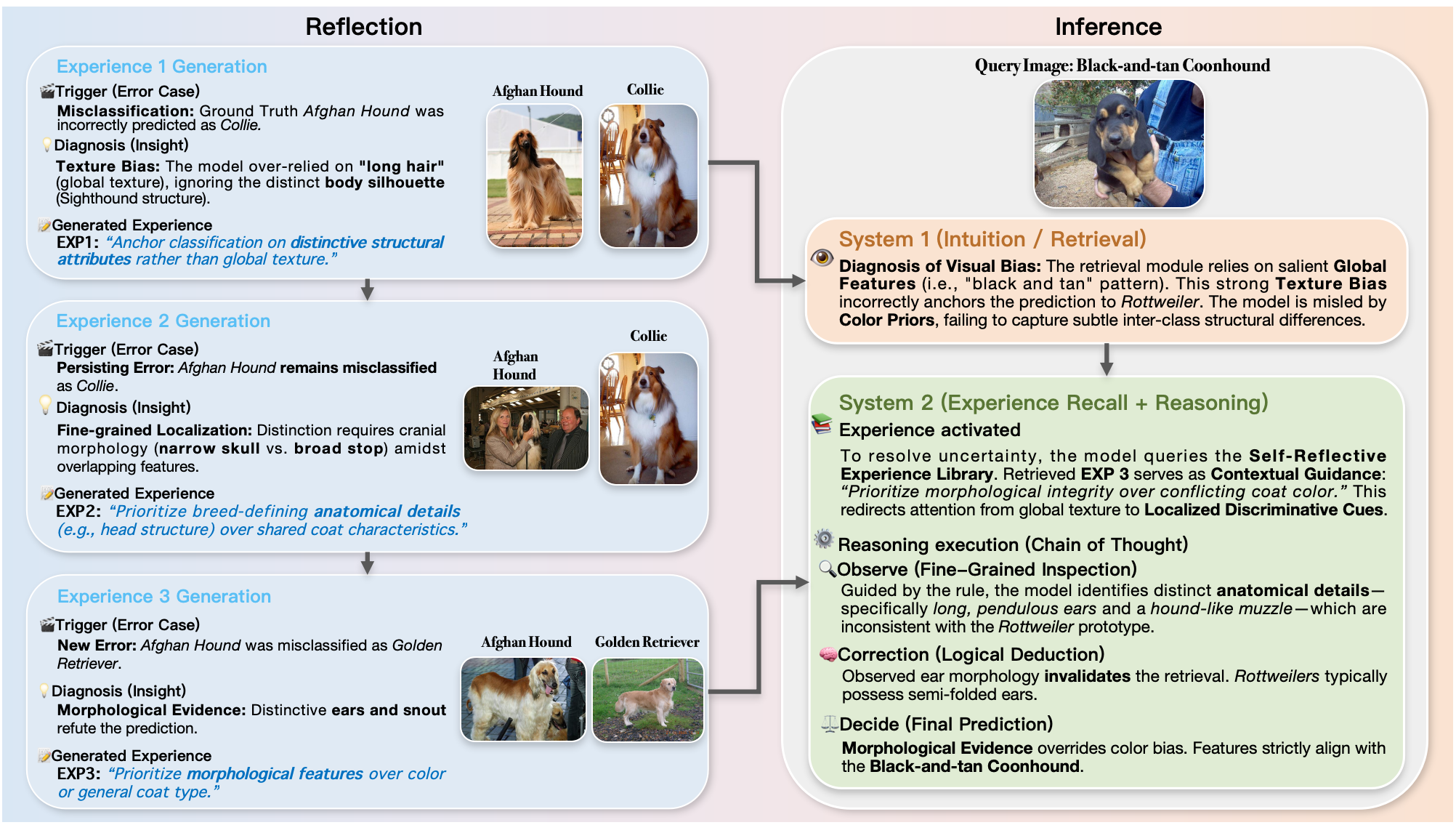}
    \caption{\textbf{Visualization of Experience Generation and Reuse.} \textbf{Left (Reflection):} The model analyzes past misclassifications (e.g., confusing an Afghan Hound with a Collie or Golden Retriever) to distill a generalized decision rule: ``Prioritize morphological features over color.'' \textbf{Right (Inference):} When encountering a visually ambiguous \textit{Black-and-tan Coonhound}, System 1 is misled by the coat color and retrieves \textit{Rottweiler}. However, System 2 retrieves the accumulated rule, corrects the focus to morphological details (ears), and successfully identifies the correct category.}
    \label{fig:case_study}
\end{figure*}

\begin{figure*}[h]
    \centering
    \includegraphics[width=\linewidth]{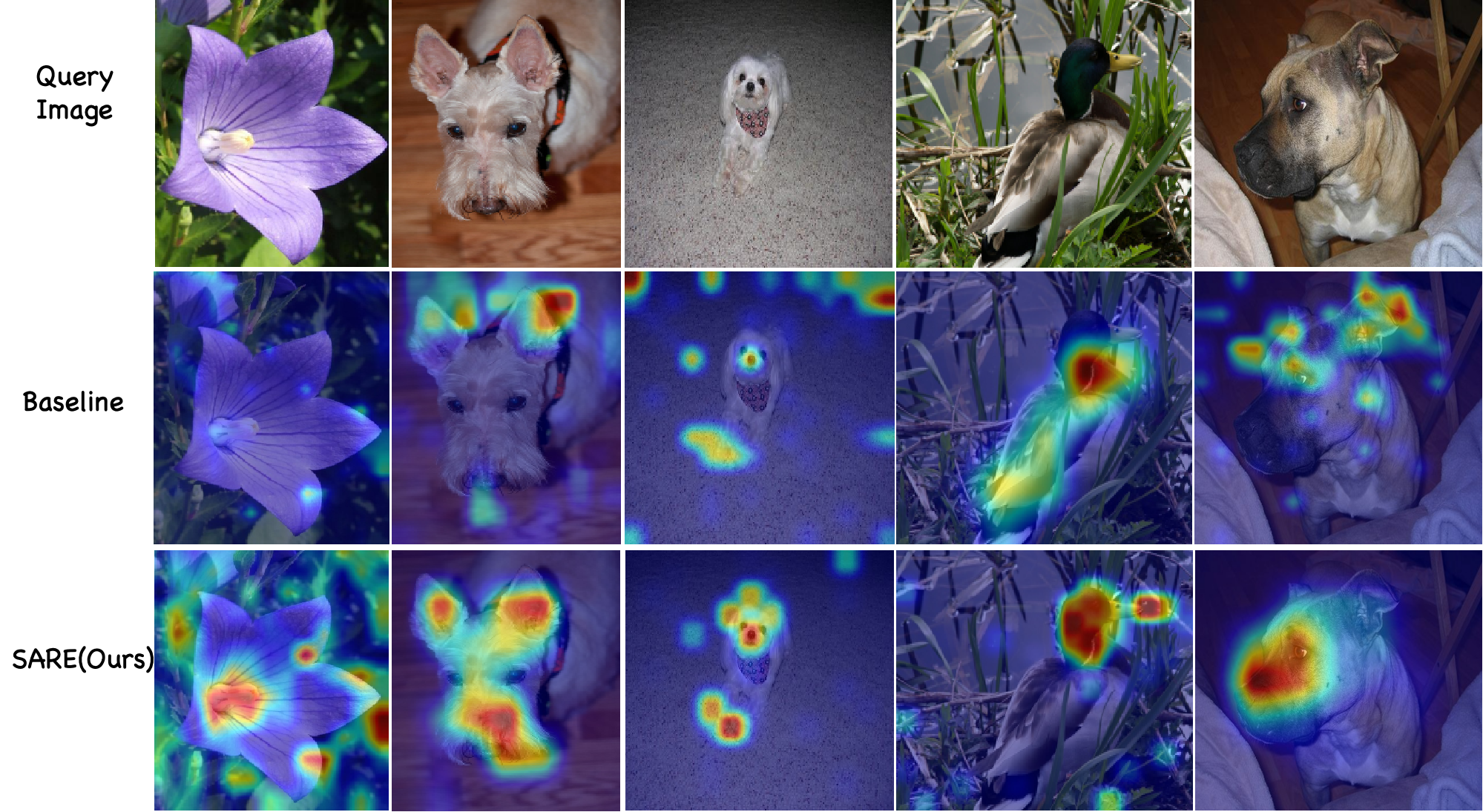}
    \caption{Visualization of attention heatmaps.}
    \label{fig:heatmap}
\end{figure*}



\section{Case Study}
\label{sec:appendix_case_study}
In this section, we provide a detailed visualization of how the Experience Library is constructed and applied to guide fine-grained recognition. As shown in Figure~\ref{fig:case_study}, the left panel illustrates the \emph{reflection phase}, where the model distills generalized discriminative rules from past misclassifications (e.g., confusing an Afghan Hound with a Collie or Golden Retriever), such as prioritize morphological features over coat color. The right panel shows the \emph{adaptive inference phase}, where a visually ambiguous sample (e.g., a Black-and-tan Coonhound) is initially misclassified by System 1 due to misleading cues. System 2 then retrieves the relevant experience rule, refocuses on critical morphological attributes (ears), and correctly identifies the category. This example demonstrates how the Experience Library captures reusable knowledge from prior errors, enabling the model to handle complex and ambiguous scenarios effectively.

\section{Related Work}
\label{sec:appendix_related_work}

Fine-Grained Visual Recognition (FGVR) has recently shifted from training-centric paradigms to training-free methods that leverage Large Vision-Language Models (LVLMs)~\cite{2019survey,intro1,peng2026survey,analysis1,performence1,performence2}.
Training-based adaptations, such as prompt learning~\cite{CoOp,CoCoOp,MaPLe} and adapter tuning~\cite{CLIP-Adapter,FineDefics,NeAR,VT-FSL,trainiclr,hogvul}, improve discrimination through domain-specific parameter updates but require substantial annotation efforts and are prone to catastrophic forgetting.
As a result, recent research emphasizes harnessing the intrinsic capabilities of frozen LVLMs, broadly categorized into retrieval-oriented perception and reasoning-oriented cognition.

\noindent\textbf{Retrieval-Oriented Paradigms.}
These methods formulate FGVR as a feature-matching or knowledge-retrieval task, prioritizing efficiency. Early works such as SuS-X-LC~\cite{sus}, FineR~\cite{FineR}, and E-FineR~\cite{E-FineR} generate hierarchical attribute descriptions to construct robust textual prototypes and bridge the modality gap.
Beyond text-based enhancement, recent approaches leverage external knowledge bases or visual caches. For example, K-Lite~\cite{KLite} and REACT~\cite{REACT} retrieve external definitions to enrich CLIP's text encoder.
To address distribution shifts and noisy samples, Tip-Adapter~\cite{TipAdapter} builds a training-free key-value cache, AWT~\cite{AWT} uses optimal transport for robust alignment, and ProtoMM~\cite{ProtoMM} and RAR~\cite{RAR} retrieve diverse visual examples to support decision-making.
Other methods, like CaSED~\cite{CaSED} and UniFGVR~\cite{UniFGVR}, relax fixed-label constraints and enrich semantic coverage, capturing intra-class variations.
Despite their efficiency, these approaches rely on static, global similarity matching and often fail on challenging samples with occlusions or subtle local differences, as they lack mechanisms for dynamic local feature alignment.

\noindent\textbf{Reasoning-Oriented Paradigms.}
These approaches exploit LVLMs' reasoning capabilities to perform in-depth analysis. AutoSEP~\cite{AutoSEP} and GLOV~\cite{GLOV} optimize prompts automatically to elicit domain-specific knowledge, while MCQA~\cite{AttentionIntervention} formulates FGVR as multi-turn Question-Answering to focus on discriminative parts.
From a representation perspective, SAV~\cite{SAV} uses sparse attention vectors from generative models as discriminative classifiers.

However, reasoning-based methods face intrinsic challenges. First, including too many candidate categories can dilute context, diverting attention from key visual cues and causing inference drift~\cite{VCD,LostInMiddle}. Second, applying heavy reasoning uniformly across all samples incurs substantial computational overhead~\cite{FrugalGPT,RouteLLM}. Finally, these methods are typically stateless, preventing accumulation of reusable experience and leading to repeated failures in similar scenarios~\cite{Reflexion}.

The most closely related methods to ours are CascadeVLM~\cite{Cascade} and SCAN~\cite{Scan}, which explore retrieval-reasoning integration. CascadeVLM adopts a cascaded framework but relies on a simple static confidence threshold to connect retrieval and reasoning, making it sensitive to uncalibrated LVLM scores~\cite{JustAsk,MultiTrust}.
SCAN endows LVLMs with System 2 reasoning for fine-grained recognition, but applies it indiscriminately, ignoring easy samples and incurring unnecessary computation~\cite{FrugalGPT,FILA}.

To address these limitations, we propose SARE, a dual-system framework with sample-adaptive triggering and a self-reflective mechanism for experience accumulation, enabling selective reasoning and improved efficiency.

\end{document}